\PassOptionsToPackage{bookmarks=false}{hyperref}
\documentclass[10pt,twocolumn,letterpaper]{article}

\usepackage{wacv}
\usepackage{times}
\usepackage{epsfig}
\usepackage{graphicx}
\usepackage{amsmath}
\usepackage{amssymb}
\usepackage{gensymb}
\usepackage{tabularx}
\usepackage{makecell}
\usepackage{tabu}
\usepackage{fancyhdr}

\usepackage[pagebackref=true,breaklinks=true,letterpaper=true,colorlinks,bookmarks=false]{hyperref} 

\wacvfinalcopy 

\def\wacvPaperID{307} 

\begin{document}
\fancypagestyle{plain}{
\fancyhf{} 
\fancyfoot[L]{Copyright IEEE 2018}
\fancyfoot[C]{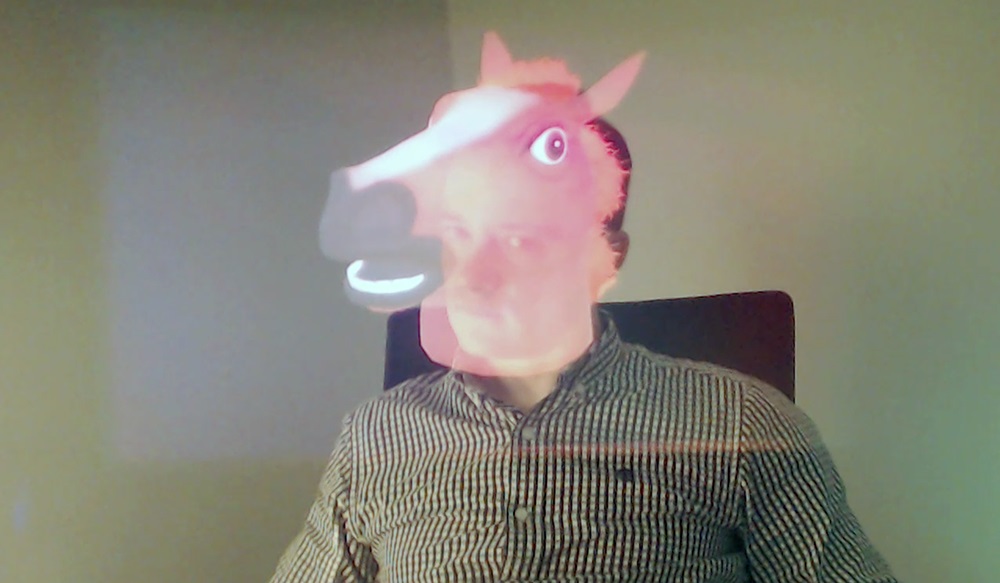}
\fancyfoot[R]{}
\fancyhead[C]{Published as a conference paper at the 2018 IEEE Winter Conf. on Applications of Computer Vision (WACV’2018)}
\renewcommand{\headrulewidth}{0pt}
\renewcommand{\footrulewidth}{0pt}
}  

\title{HoloFace: Augmenting Human-to-Human Interactions on HoloLens}

\author{Marek Kowalski, Zbigniew Nasarzewski, Grzegorz Galinski, and Piotr Garbat \\
\\
Warsaw University of Technology, Poland\\
{\tt\small \{m.kowalski,z.nasarzewski,g.galinski\}@ire.pw.edu.pl, p.garbat@elka.pw.edu.pl}
}

\maketitle

\begin{abstract}
We present HoloFace, an open-source framework for face alignment, head pose estimation and facial attribute retrieval for Microsoft HoloLens. HoloFace implements two state-of-the-art face alignment methods which can be used interchangeably: one running locally and one running on a remote backend. Head pose estimation is accomplished by fitting a deformable 3D model to the landmarks localized using face alignment. The head pose provides both the rotation of the head and a position in the world space. The parameters of the fitted 3D face model provide estimates of facial attributes such as mouth opening or smile.
Together the above information can be used to augment the faces of people seen by the HoloLens user, and thus their interaction.
Potential usage scenarios include facial recognition, emotion recognition, eye gaze tracking and many others. We demonstrate the capabilities of our framework by augmenting the faces of people seen through the HoloLens with various objects and animations. 
\end{abstract}

\begin{figure*}
\centering
\includegraphics[width=0.32\textwidth]{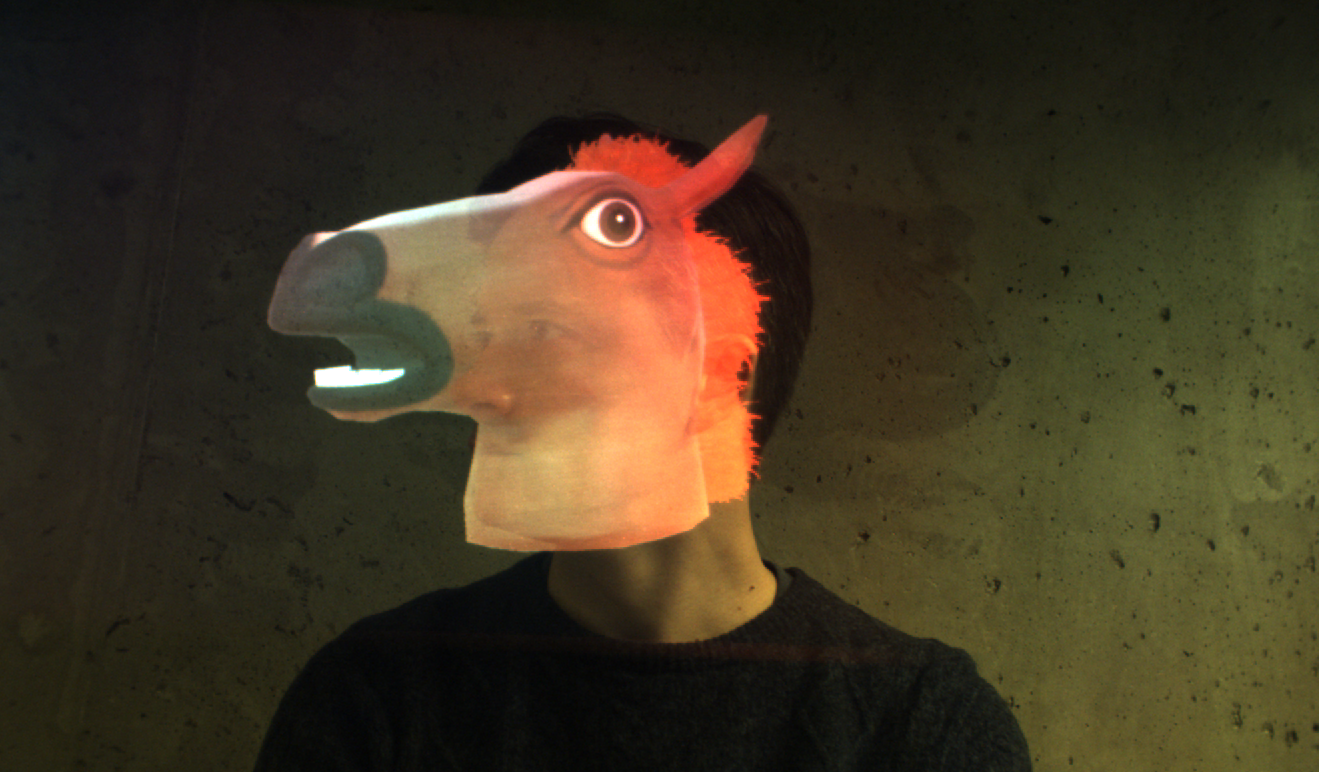}
\includegraphics[width=0.32\textwidth]{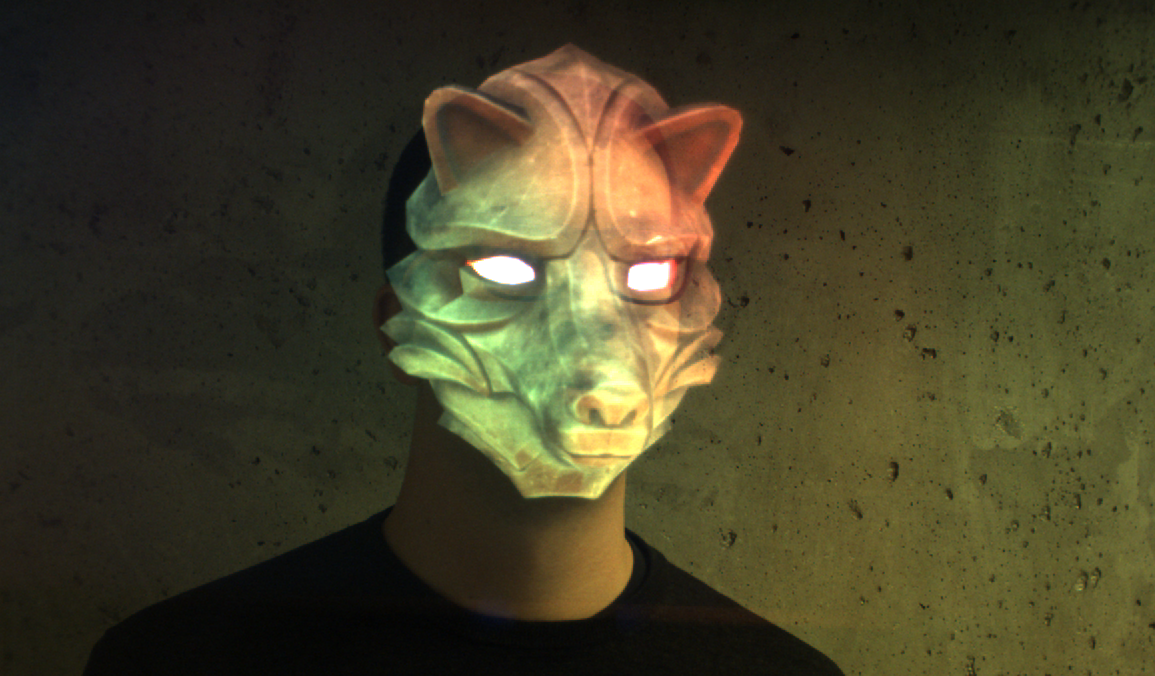}
\includegraphics[width=0.32\textwidth]{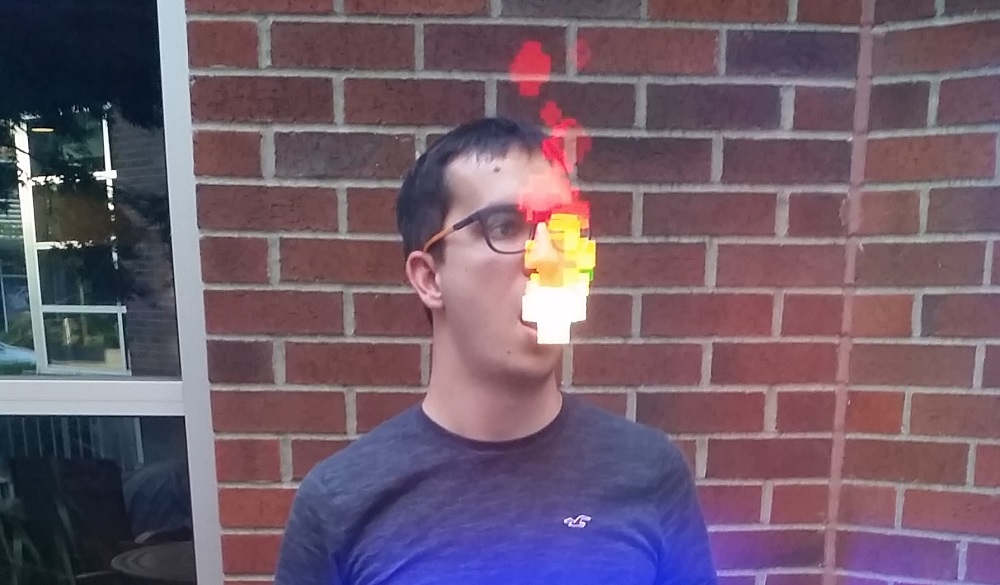}
\caption{Examples of faces augmented with animations and overlayed objects using HoloFace. Photographs taken directly through the display of Microsoft HoloLens.}
 \label{fig:example}
\end{figure*}

\section{Introduction}
With the recent release of Microsoft HoloLens, a milestone in Augmented Reality headsets, and prospective releases of similar devices from companies such as Magic Leap and Meta, the possibilities of AR technology are becoming more realizable. However, none of the proposed headsets have native capabilities that allow augmenting human-to-human interactions.

With HoloFace, we want to introduce an open-source framework that allows Augmented Reality (AR) developers to localize human faces in 3D and estimate their attributes. This can in turn be used to augment the interaction between the user of the AR headset and other people.

Potential applications range from entertainment, where HoloFace can be used to augment faces seen by the user with 3D models of game characters, to medicine, where HoloFace can be combined with face recognition to automatically show data about patients seen by the doctor. Figure \ref{fig:example} shows an example of HoloFace with the subjects' faces augmented with items and visual effects.

HoloFace uses the frontal camera of an AR headset to first detect and later track the subject's face and its landmarks. The landmarks are subsequently used to fit the CANDIDE-3 \cite{candide} deformable 3D face model to the detected face. The pose of the fitted model,  which consists of translation and rotation, is converted to the pose of the subject's face in the world coordinate system. This allows for rendering items on, or around the subject's face. The weights of the blendshapes of the fitted model are used to estimate the attributes of the subject's face, such as whether the subject is smiling, opening his mouth etc.

Since HoloLens is a battery powered device, performance is a key factor, all of the methods that run on the device are chosen with performance in mind. This approach leads to trade-offs in speed vs accuracy. In order to reduce this problem we include two face tracking methods, one that is optimized for speed and runs on the device, and one that is more accurate and runs on a remote machine.

HoloFace is implemented using the Unity game engine, with the most computationally expensive elements implemented in C++ as plugins.  It is important to note that while currently HoloFace only works with Microsoft HoloLens, it should be easily portable to other Windows Mixed Reality Devices (provided they have a frontal camera), once they become available.

The main contributions of the article are the following:
\begin{itemize}
  \item a design of a framework that allows for face alignment and 3D head pose estimation on a low-power device such as Microsoft HoloLens,
  \item a simple and effective method for verifying face tracking failure in neural network based face alignment methods,
  \item open-source implementation of the proposed framework.
\end{itemize}

The remainder of this paper is organized as follows: section \ref{sec:related} reviews the related work, section \ref{sec:methods} details the methods we have used, section \ref{sec:implementation} describes the implementation and section \ref{sec:experiments} explains the measurement of the HoloLens camera latency and verifies the accuracy of the face tracking and failure detection methods.

\section{Related work}\label{sec:related}
This article is based on previous work in several areas including: augmented reality, face alignment and head pose estimation. For brevity we will only discuss related work in the area of augmenting facial images which is the main topic of this paper. 

The augmentation of facial images is a popular application of AR in industry. In the consumer sector, Snapchat and MSQRD have proposed mobile applications that render animations and objects on the image of a user's face. In the professional sector, FaceRig, Adobe and others offer applications that allow for markerless facial motion capture and automated animation of character models. 

Astonishingly, augmenting facial images on mobile devices has received relatively little attention in academia. One of the first works that touched on this topic was \cite{Dantone11}, where the authors propose a pipeline that uses face tracking and facial recognition on a smartphone. The application augments the faces of people seen through the smartphone's camera with information about them based on their recognized identity. 

One of the downsides of the method proposed in \cite{Dantone11} is that the pose of the head is not estimated, which means that the face can only be augmented with flat objects. In contrast, \cite{kitanovski2011} tracks the 3D positions of facial landmarks which allows for augmenting the face image with 3D effects. More recently, \cite{Wang2017} proposed a method in which 2D landmarks are tracked and the head pose is estimated using a Perspective-n-Point (PnP) method which also allows for rendering 3D effects on the user's face. In \cite{MakeupLamps} the authors propose quite a different facial augmentation system, that displays the augmented content directly on the subject's face using a projector. The use of specialized hardware allows for very high framerates and visually appealing effects, on the downside the augmented content can only be displayed on the face itself, not outside of it. Neither of \cite{kitanovski2011, Wang2017, MakeupLamps} are intended for mobile platforms.

HoloFace provides the locations of 2D and 3D landmarks (the latter in the form of the vertices of the fitted 3D model) as well as the head pose and facial attributes. Thanks to the nature of the HoloLens display it also allows for displaying content both on and around the subject's face. Moreover, to the best of our knowledge no other work has previously covered the problem of augmenting faces, and consequently interactions, on Augmented Reality headsets.

\begin{figure*}
\centering
\includegraphics[width=0.9\textwidth]{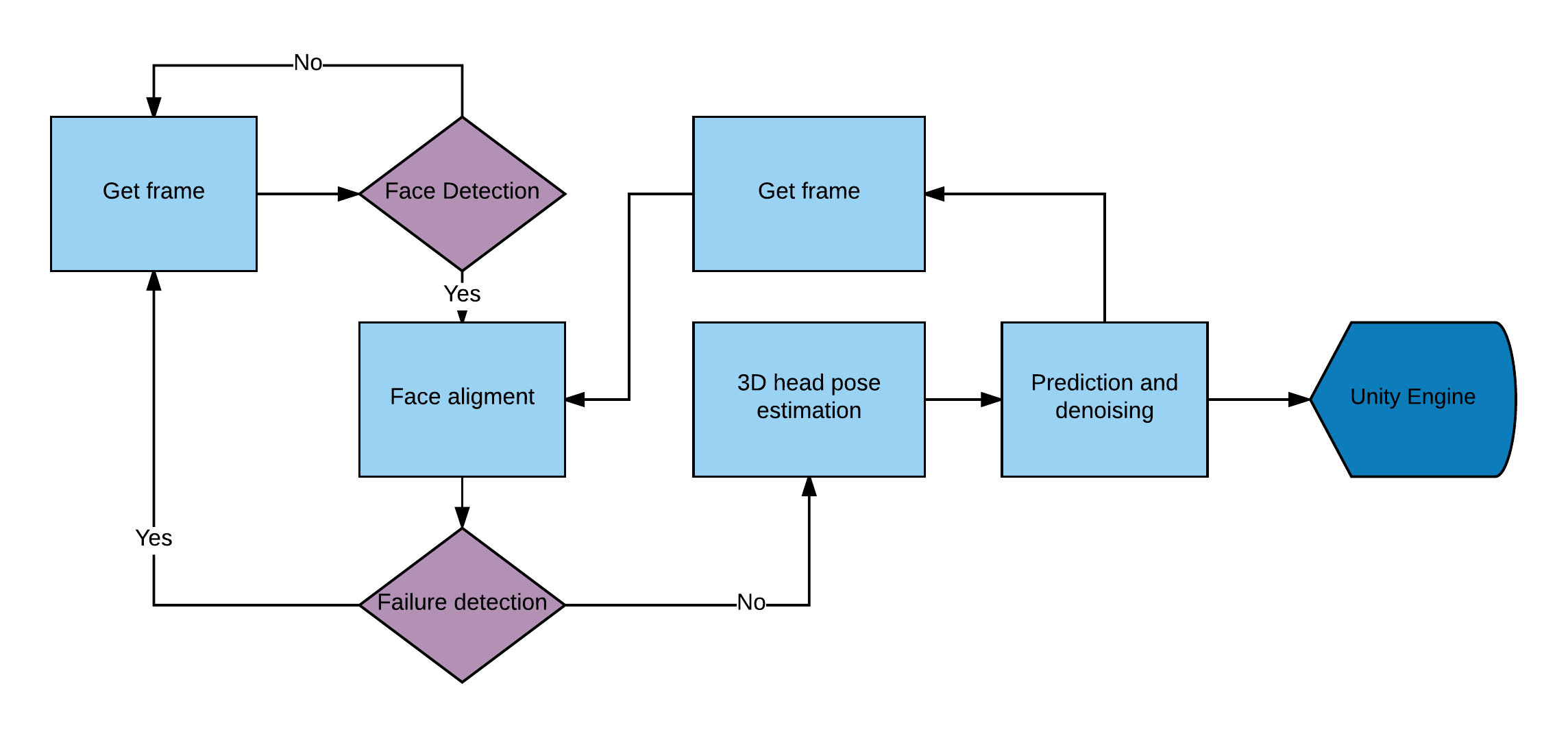}
\caption{A diagram showing the outline of the HoloFace framework.}
 \label{fig:diagram}
\end{figure*}

\section{Overview}\label{sec:overview}
Figure \ref{fig:diagram} shows an outline of the HoloFace framework. In the initial state, when no face is being tracked the framework performs face detection on the arriving frames. Once a face is detected its landmarks are localized using one of the two face alignment methods included in the pipeline (details in section \ref{sec:alignment}). In the subsequent frames the face is tracked along with its landmark using the same face alignment method. During tracking the face alignment is initialized based on the landmark locations in the previous frame, taking into account the headset movement (details in section \ref{sec:tracking}).

For each frame where the face is tracked the quality of the tracking is verified. Here we again use two different methods which are specific to the two face alignment methods, more details in section \ref{sec:tracking}. If the quality of the tracking is less than a specified threshold the tracking is considered to have failed and the pipeline goes back to the face detection step.

If the landmarks have been located successfully the 3D head pose and the facial attributes are estimated using the methods described in sections \ref{sec:headpose} and \ref{sec:attributes}. This step is followed by the denoising and prediction step which aims to both reduce the jitter and latency of the head pose. The prediction is necessary to compensate for the delay between image acquisition and rendering, more details in section \ref{sec:prediction}. 

The final head pose is passed to the Unity Engine which can use it to render any object on or around the subject's face.

\section{Methods}\label{sec:methods}
For the face detection step of the HoloFace framework we use the face detection method built into the Windows Universal Platform framework, the remaining methods are described below
.
\subsection{Face alignment}\label{sec:alignment}
Many of the recently proposed face alignment methods are based on deep neural networks \cite{DAN, Hourglass, PoseInvariantICCV2017}. Unfortunately the computational complexity of deep neural networks limits their application on mobile devices. Most recently the authors of \cite{Binarized} have proposed a method for face alignment based on a binarized convolutional neural network which has a potential to run on a mobile device, however no test on an actual device were performed to validate this capability. On the other hand face alignment methods based on more traditional approaches such as regression trees \cite{KRFWS, LBF, Kazemi} are fast enough to run on a mobile device, but offer less accurate results.

Face alignment is the most crucial part of our framework that all the other elements are based on. For that reason in HoloFace we implement two state-of-the-art face alignment methods: a method based on regression trees which is capable of running locally on the device \cite{KRFWS} and a more powerful method based on deep neural networks which is intended to run on a remote desktop machine \cite{DAN}.

The first method, which we will refer to as KRFWS, is based on the work of Kowalski et al. in \cite{KRFWS}. The authors of \cite{KRFWS} propose a face alignment pipeline that uses novel K-Cluster Regression Forests with Weighted Splitting. 

In HoloFace, KRFWS is intended to perform all of the image processing locally on the HoloLens. Because of that, we had to simplify the method to achieve reasonable processing speed. To that end we only use the base face alignment method without initialization refinement (called APR and 3D-APR in the original article). 
Moreover, we substitute the Pyramid Histogram of Oriented Gradients (PHOG) \cite{KRF} features with standard Histogram of Oriented Gradients (HOG) \cite{HOG} features and reduce their size to $16\times 16$ pixels per landmark. We also reduce the amount of face alignment stages to 3.

\begin{figure}
\centering
\includegraphics[width=0.49\linewidth]{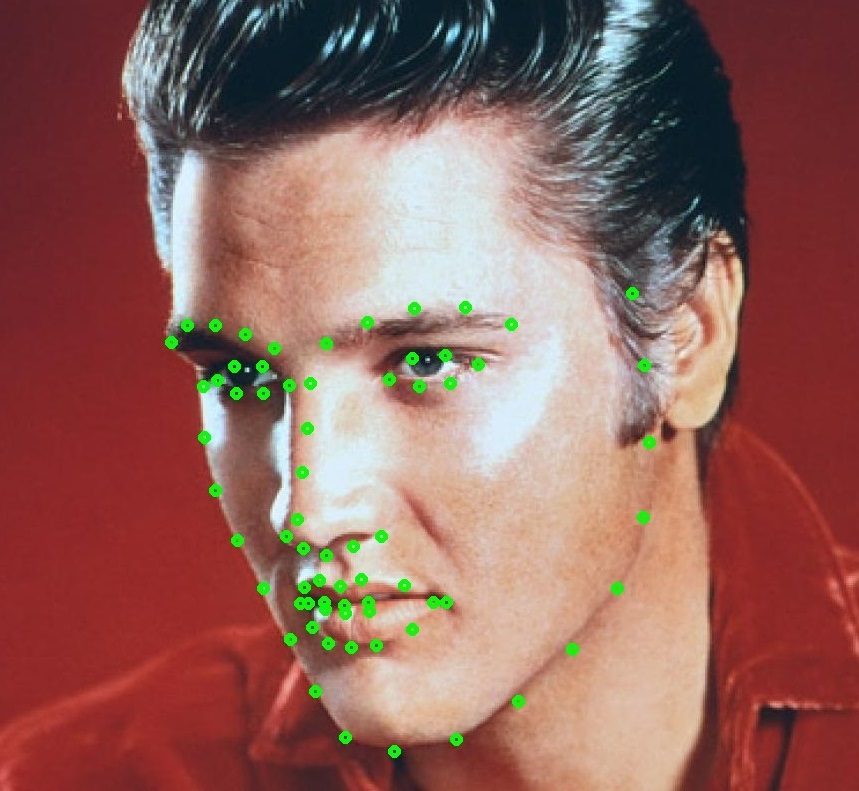}
\includegraphics[width=0.49\linewidth]{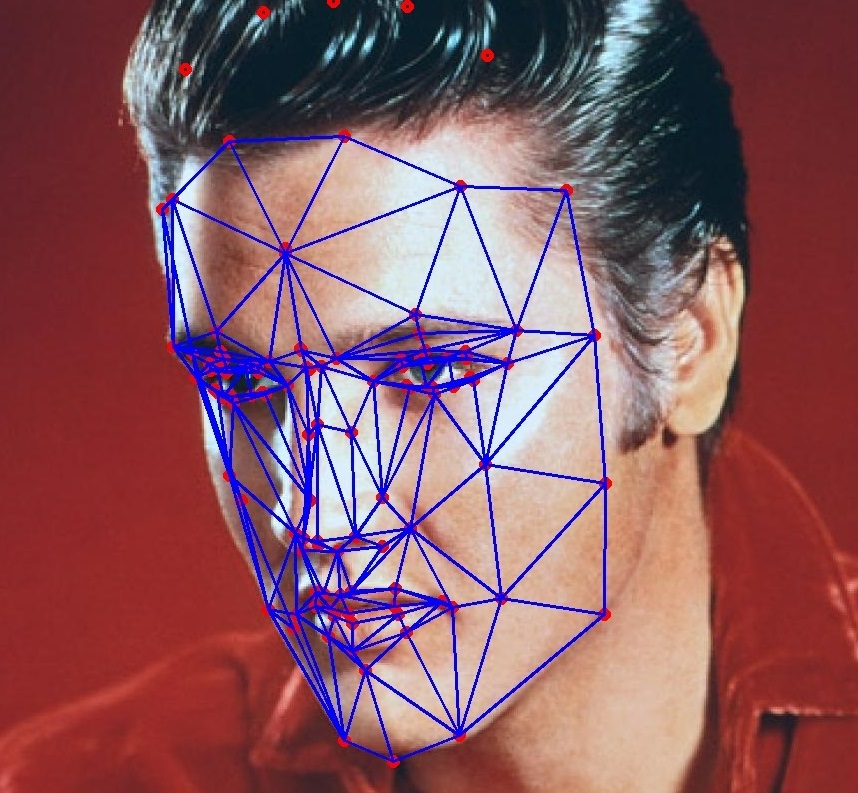}\\
\includegraphics[width=0.49\linewidth]{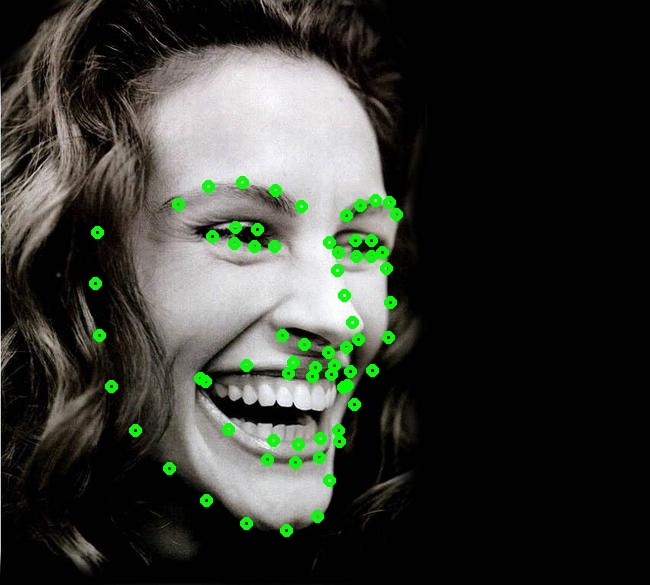}
\includegraphics[width=0.49\linewidth]{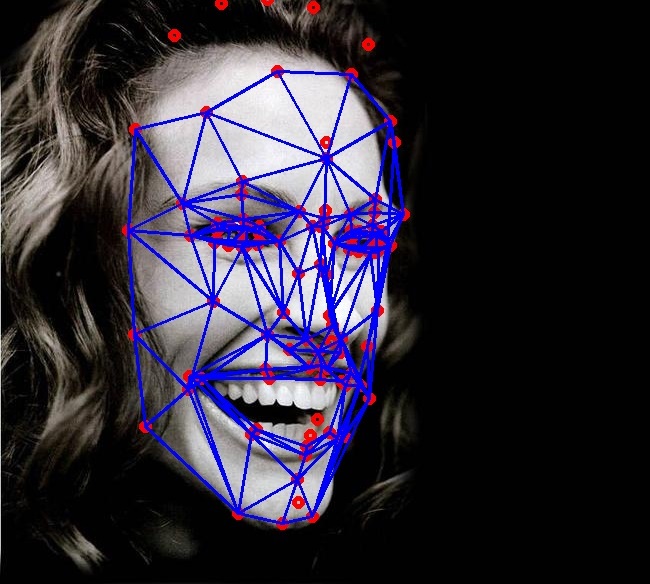}\\
\includegraphics[width=0.49\linewidth]{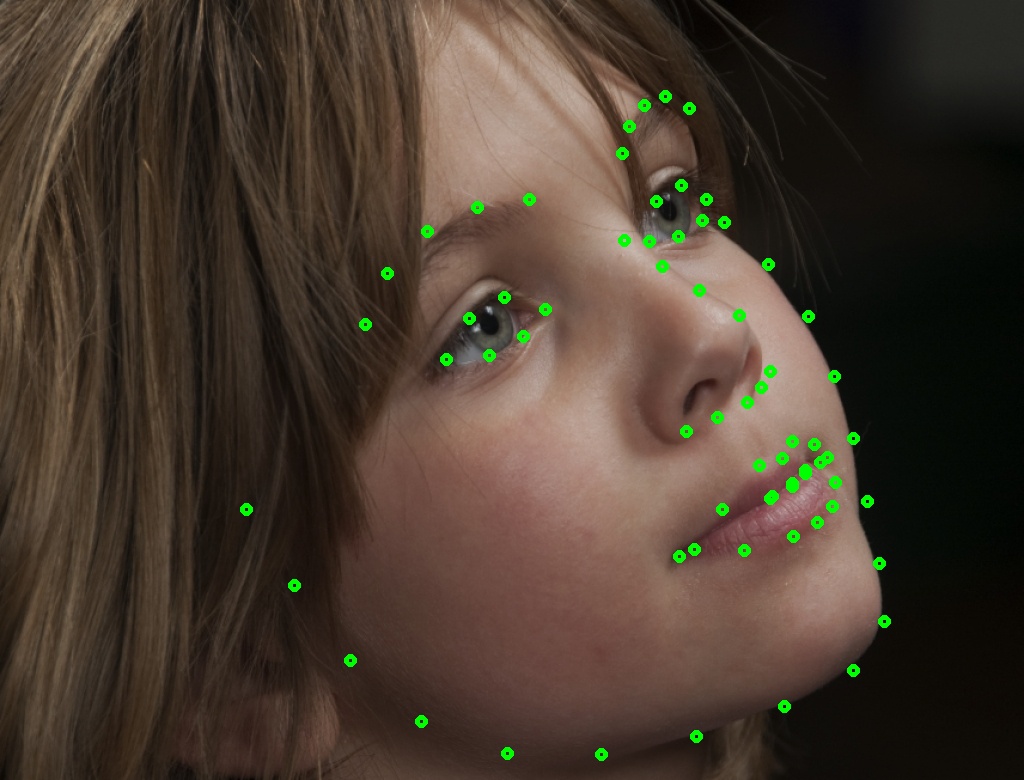}
\includegraphics[width=0.49\linewidth]{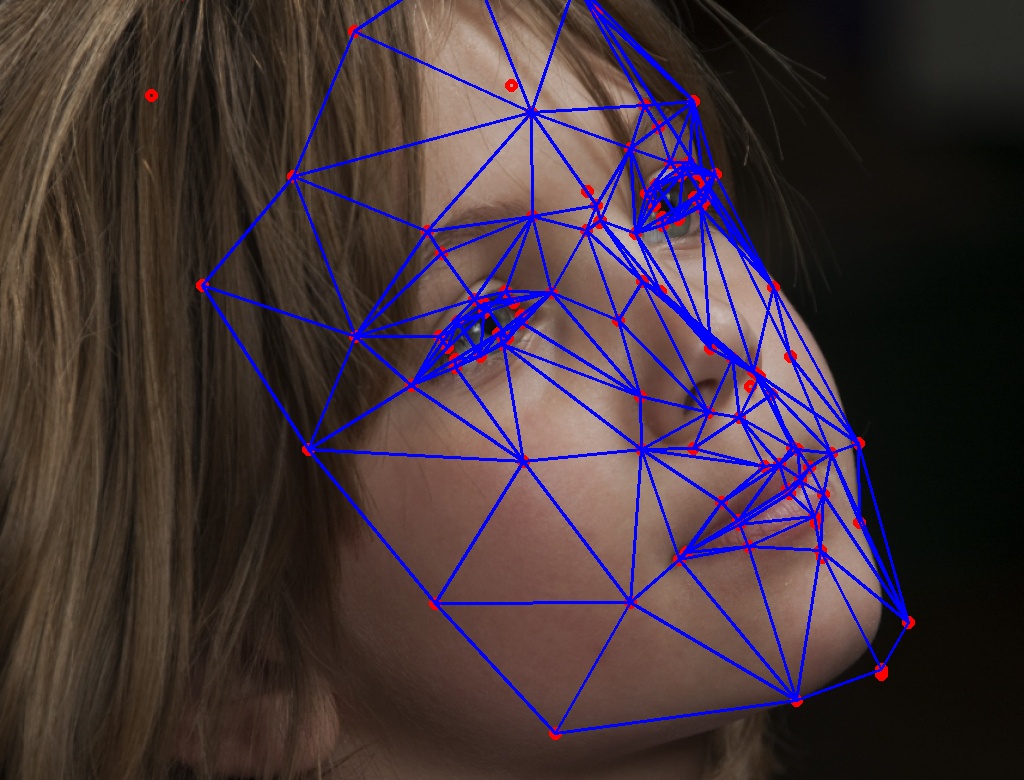}\\
\caption{Images from the 300-W dataset \cite{300W} with landmarks localized using DAN on the left and with the fitted CANDIDE-3 model on the right.}
 \label{fig:landmarks}
\end{figure}

The second, alternative, face alignment method we use is Deep Alignment Network (DAN) \cite{DAN}. Since this recently published method uses convolutional neural networks, it is too expensive to execute it locally on the HoloLens. Because of that, DAN is run on a remote computer equipped with a CUDA enabled GPU for fast processing. The communication between the AR headset and the computer takes place over WiFi. In order to increase the processing speed of DAN we only use a single stage of the neural network, as opposed to two stages employed in the original article. Examples of images with landmarks localized using DAN are shown in Figure \ref{fig:landmarks}.

There are many situations where the two approaches can be used complimentarily. For example, if an application based on HoloFace is mostly used within a given building it can use the more accurate remote backend while on WiFi, and easily switch to the local tracker when outside of range. 

Section \ref{sec:experiments} contains a comparison of both methods in terms of tracking accuracy.

\subsection{Head pose estimation} \label{sec:headpose}
Since performance is a key factor in HoloFace, we propose a simple, fast, optimization based head pose estimation method. The pose of the head is estimated by fitting the CANDIDE-3 \cite{candide} deformable 3D face model to the localized landmarks. The CANDIDE-3 model consists of an average 3D face shape $S_0$ as well as a number of shape units and action units, which can be used to deform $S_0$. We refer to the shape and action units together as blendshapes and denote them with $S_{1..n}$ where $n$ is the total number of blendshapes. The mean shape $S_0$ as well as the blendshapes consist of 113 vertices connected with 168 triangles. 

The fitting is accomplished by solving the problem below using the Gauss-Newton method:
\begin{gather}
\min_{R, t, w_{1..n}} \left\Vert s - proj\left( K R \left( S_0 + \sum_{i=1}^{n} w_i S_i \right) + Kt  \right) \right\Vert ^ 2, \\ 
proj\left( 
\begin{bmatrix} 
x\\
y\\
\lambda \\
\end{bmatrix}
\right) = 
\begin{bmatrix}
\frac{x}{\lambda} \\[6pt]
\frac{y}{\lambda} \\
\end{bmatrix},
\end{gather}
where $s$ are the localized landmarks, $K$ is the intrinsic matrix of the camera mounted on the AR headset, $[R,t]$ are the rotation matrix and translation vector describing the head pose and $w_{1..n}$ are the blendshape weights. Since the set of landmarks localized using face alignment and the set of vertices in CANDIDE-3 might be different, we only use a manually selected subset of landmarks that are common between the two sets.

In order for the obtained pose to be metric we scale $S_{0..n}$ so that the inter-pupillary distance of $S_0$ is equal to that of the average person - 63mm \cite{IPD}.

The head pose $[R,t]$ obtained from optimization is in the coordinate system of the camera located on the AR headset. In order to render objects on the face we need to obtain its pose $[R_w, t_w]$ in the world coordinate system. To do so it is necessary to know the camera to world transform $[R_{cw}, t_{cw}]$ of the headset at the time the image of the face was taken. This transform is easily obtained from the HoloLensForCV API \cite{HoloLensForCV}.

Given that $[R_{cw}, t_{cw}]$ is known, the position of the head in the world coordinate system is calculated as follows:
\begin{gather}
t_w = R_{cw}t + t_{cw} \\
R_{la} = lookAt(t_w, t_{cw}) \\
R_w = R R_{la},
\end{gather}
where $lookAt$ outputs a rotation that points the forward vector from $t_{w}$ to $t_{cw}$ (the camera's position in the world coordinate system) . 

Several examples of images with the mesh of the fitted CANDIDE-3 face model overlaid on them are shown in Figure \ref{fig:landmarks}.

\subsection{Facial attribute estimation} \label{sec:attributes}
The blendshape weights $w_{1..n}$ of the fitted model are used to estimate the facial attributes of the tracked face. If the weight of a given blendshape exceeds a predefined threshold the corresponding attribute is considered to be present. This can be used to trigger animations, recognize emotions, etc. The facial attributes we can recognize with this method are the following: smiling, eyebrow raising, mouth opening.

\subsection{Face tracking} \label{sec:tracking}
In most face tracking applications the camera is stationary and for each new frame the tracker is initialized with the facial landmarks from the previous frame. On AR devices the camera is moving together with the user's head, which can lead to very large differences between the location of the face in the image in consecutive frames.

In order to compensate for the movement of the user's head we take advantage of the fact that the pose of the HoloLens is known at every moment in time. We use the previously estimated head pose to obtain the 3D world space coordinates $S^{prev}$ of the facial landmarks $s^{prev}$ from the previous frame. We then project them into the image space using the world-to-camera transform $[R_{wc}, t_{wc}]$ of the current frame to obtain the initial landmarks for the tracker $s^{init}$. 
\begin{gather}
S_{unproj} = K^{-1}s^{prev},\\
S^{prev} = t_{cw}^{prev} + \frac{R_{cw}^{prev} S_{unproj}}{\Vert S_{unproj} \Vert} \cdot \Vert t_{cw}^{prev} - t_w \Vert, \\
s^{init} = proj \left( K \left( R_{wc}S^{prev} + t_{wc} \right) \right),
\end{gather}
where $[R_{cw}^{prev}, t_{cw}^{prev}]$ is the camera to world transform from the previous frame. It is important to note that $t_{cw}^{prev}$ is not the same as $t_{cw}$ for the current frame as the headset might have changed its position between the moments the frames were taken. The world-to-camera and camera-to-world transform are obtained from the HoloLensForCV API \cite{HoloLensForCV}.

As a result of the procedure described above, the initialization of the face tracker is invariant to the head movement of the headset user. This greatly decreases the number of times loss of tracking occurs and thus makes the experiences more stable.

It is important to note that loss of tracking may still occur, for example, if the subject's face moves very quickly or becomes occluded. Because of that it is necessary to detect it and reinitialize the tracker if it occurs. In HoloFace both face alignment methods (KRFWS and DAN, see section \ref{sec:alignment} for details) have their own methods to detect loss of tracking. The decision to use different methods is motivated by the fact that the two face alignment algorithms are based on very different principles.

In KRFWS we detect loss of tracking using a simple method based on the Supervised Descent Method \cite{SDM}. In this method we first extract HOG features in patches around each localized landmark. The HOG features are subsequently concatenated and passed to a linear model which predicts the Sum of Squared Errors of the landmark positions. If the predicted error exceeds a predefined threshold the tracking is considered to have been lost. In order to save computational resources we use the HOG features extracted in the last stage of KRFWS instead of extracting new features. Thanks to this approach the loss of tracking detection adds very little overhead.

In DAN the loss of tracking is detected using an additional layer added after the penultimate layer of the original network. The layer consists of two neurons densely connected to the previous layer, followed by a softmax nonlinearity. This layer is trained to recognize whether the input images contains a face or not. Thanks to this approach we can perform training on very large datasets designed for face detection rather than on smaller datasets that have annotated landmark locations. The resulting method is capable of detecting loss of tracking and adds almost no additional computational overhead. It is important to note that this simple method can be easily added to nearly any neural network based face alignment method. 

\begin{figure}
\centering
\includegraphics[width=0.49\linewidth]{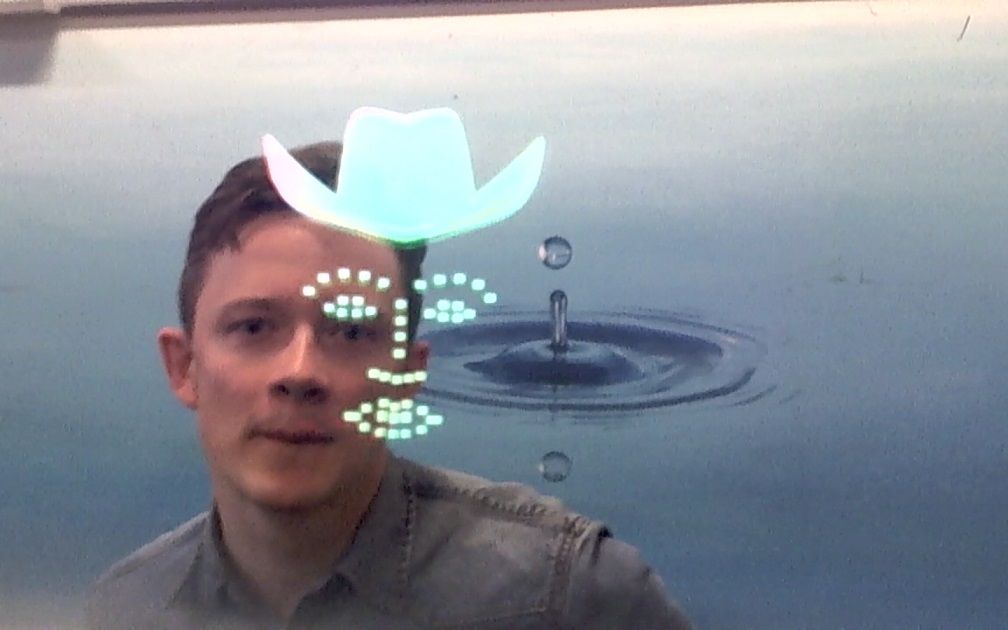}
\includegraphics[width=0.49\linewidth]{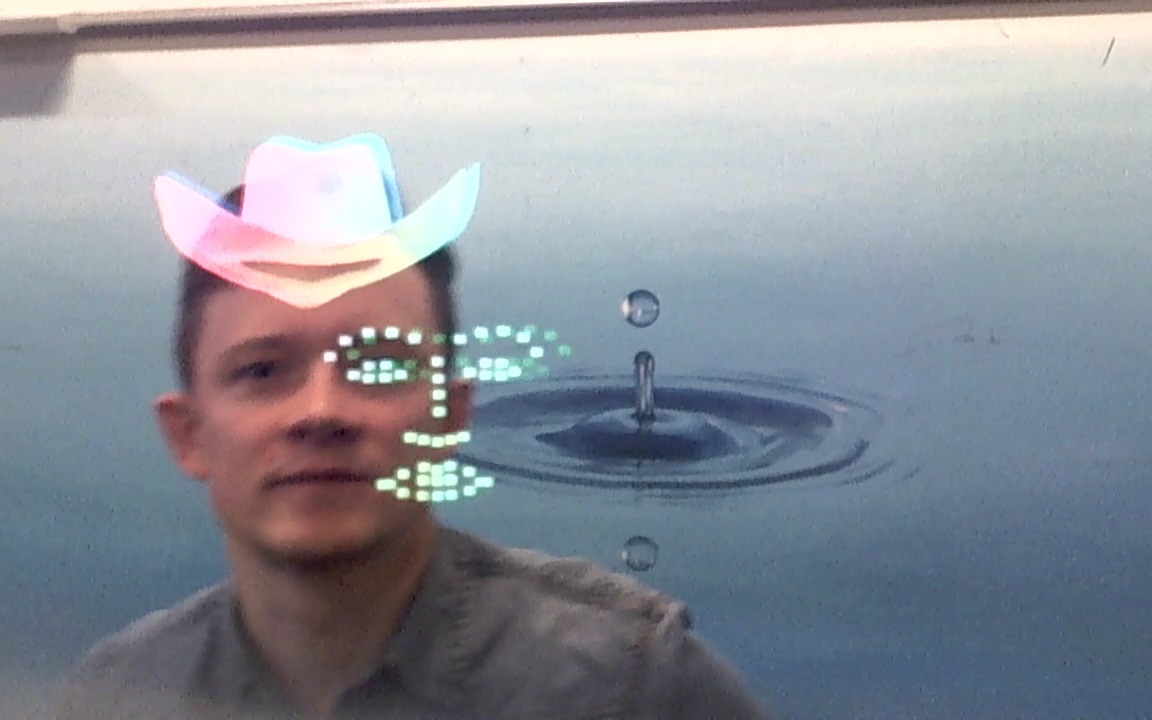}\\
\caption{Comparison of HoloFace with and without the prediction step, both images are a part of a sequence where the subject's head is moving from right to left. In the image on the left no prediction is performed and the object clearly lags behind the face, in the image on the right the proposed prediction scheme reduces the delay. The rendered landmarks indicate where the face was at the time of image acquisition. }
 \label{fig:prediction}
\end{figure}

\subsection{Prediction and denoising} \label{sec:prediction}
One of the key differences between augmenting facial images through a headset and augmenting them on a screen (as in popular smartphone applications), is the influence of latency. In the latter case even if the time between image acquisition and display is significant, the perceived latency is small because all of the augmentation is shown rendered on top of the original image. In case of a headset the user sees the scene and the objects rendered on top of it with his own eyes. This makes any delay between the image acquisition and rendering visible. 

For example, if the subject's head is moving then any objects rendered on top of that head will drag behind. This is caused by the fact that the pose of the rendered objects is based on where the subject's head was at the time of image acquisition and in the time between acquisition and rendering the head has moved, for an example see Figure \ref{fig:prediction}. In order to reduce this effect we employ a Kalman filter \cite{Kalman} to predict the location of the subject's head at rendering time. 

The formulation of the Kalman filter we use is similar to the one proposed in \cite{MakeupLamps}. The position of the head is modeled by three Kalman filters, one for each axis. The state vector $(x, v, a)$ consists of the position along the given axis $x$, velocity $v$ and acceleration $a$. The only measured quantity is the position. The process transition $A$, process noise covariance $Q$ and measurement noise covariance $R$ matrices are defined as follows: 
\begin{equation}
A = 
\begin{pmatrix}
1 & \Delta t & \frac{\Delta t^2}{2} \\
0 & 1 & \Delta t \\
0 & 0 & \alpha 
\end{pmatrix},
Q = 
\begin{pmatrix}
0 & 0 & 0 \\
0 & 0 & 0 \\
0 & 0 & q^2_m 
\end{pmatrix},
R = 
\begin{pmatrix}
\sigma_z^2
\end{pmatrix},
\end{equation}
where $\Delta t$ is the time step, $\alpha = e^{\frac{-\Delta t}{\tau_m}}$, $q_m = \sigma_a (1- \alpha)^2$, $\sigma_z$ and $\sigma_a$ are the standard deviations of the acceleration and measurement noise, while $\tau_m$ is the decay rate to white noise. Through testing on a prerecorded sequence we determine the optimal values of the parameters to be $\sigma_a = 10$, $\sigma_z = 0.01$, $\tau_m = 0.1$. At those values the filters achieve the most accurate prediction of the face localization in 120ms. In testing we have found out that predicting the location of the face further into the future leads to a larger error, even if the actual delay is higher. This is caused by the predictions overshooting the correct position when the head changes direction.

At runtime $\Delta t$ is set to the time that elapsed since last measurement. The localization of the face that is passed to Unity Engine for rendering is predicted based on the time elapsed since image acquisition.

The Kalman filter also serves an additional purpose of reducing the influence of the noise in landmark localization on the head pose. In order to further reduce this influence we employ an additional filtering step in which we average the head pose over the two most recent frames if the head moved by less than 5mm and turned by less than $4\degree$.


\section{Implementation}\label{sec:implementation}
\begin{figure*}
\centering
\includegraphics[width=0.24\linewidth]{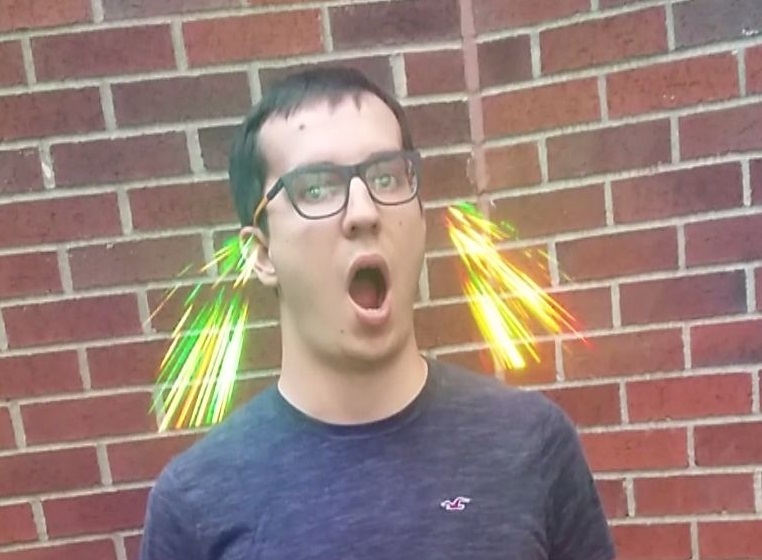}
\includegraphics[width=0.24\linewidth]{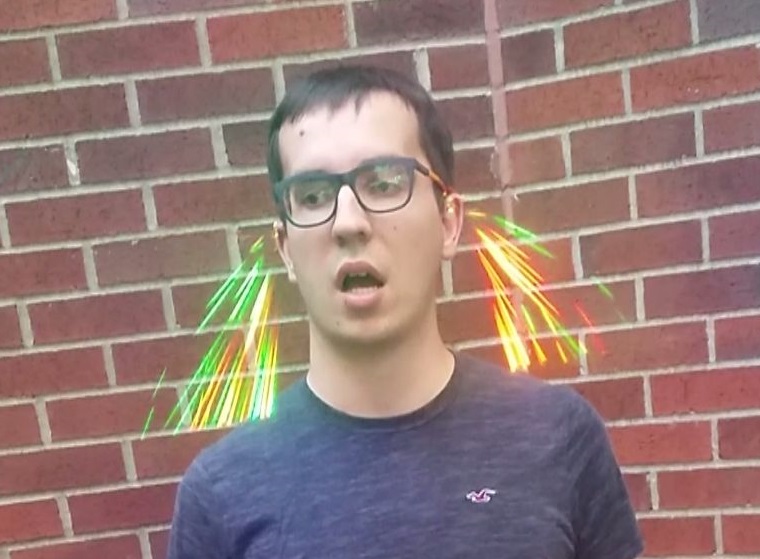}
\includegraphics[width=0.24\linewidth]{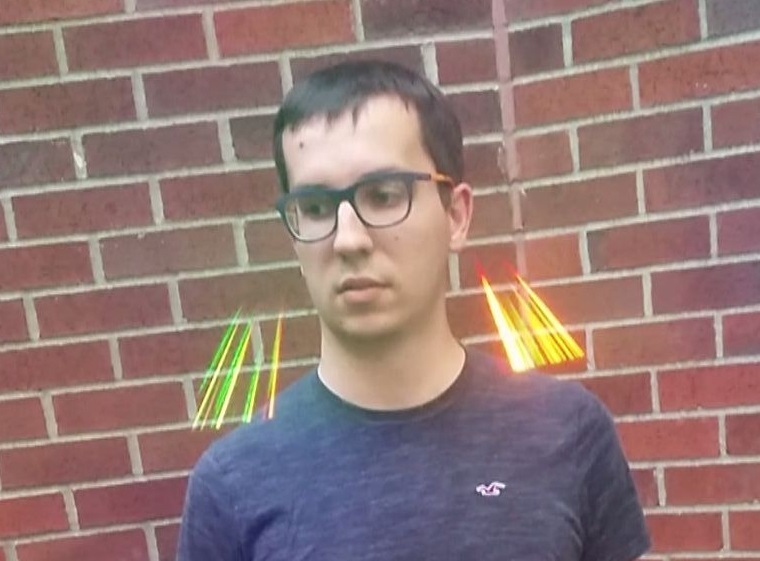}
\includegraphics[width=0.24\linewidth]{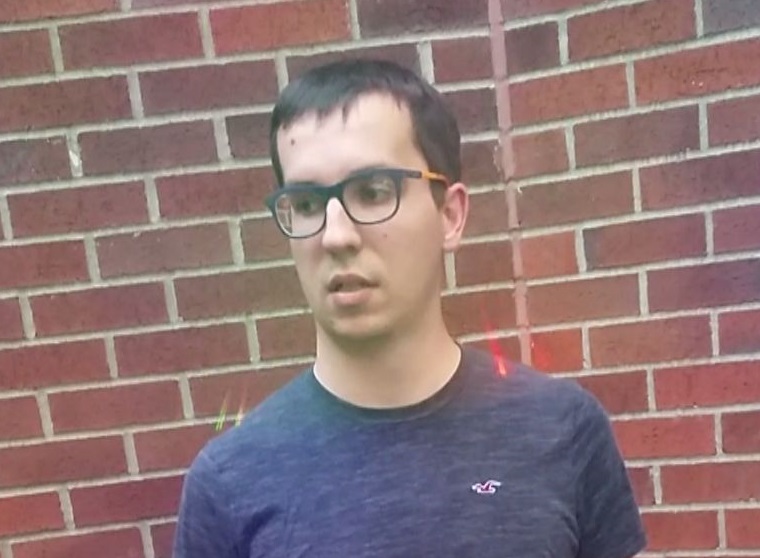}
\caption{Several frames from a sequence showing a face augmented with sparks flying out of the ears, the animation is enabled when the subject opens his mouth. Images taken directly through the display of Microsoft HoloLens.}
 \label{fig:sequence}
\end{figure*}

HoloFace is implemented in the Unity game engine with the most compute intensive elements, such as local face tracking, implemented in a separate C++ plugin. The use of Unity allows for easy integration of HoloFace with existing software and easy repurposing to other applications. 

Please note that, except for the remote facial landmark tracking method, all of the elements of HoloFace run locally on the device. The remote facial landmark tracking method, once enabled, runs for as long as the connection with the server is maintained and switches to the local method when connection is lost. To summarize, if there is no connection to the backend processing server, HoloFace runs entirely on the device.

In our implementation we present an application of HoloFace to entertainment which allows for rendering objects and animations on top of a face seen through the headset. The application is controlled using gestures and voice commands which allow the user to choose the effect being displayed, enable the remote tracker and enter a debug mode which shows the framerate and landmark locations. Thanks to the Unity Engine new objects and animations can be easily added or edited. 

\begin{figure}
\centering
\includegraphics[width=0.75\linewidth]{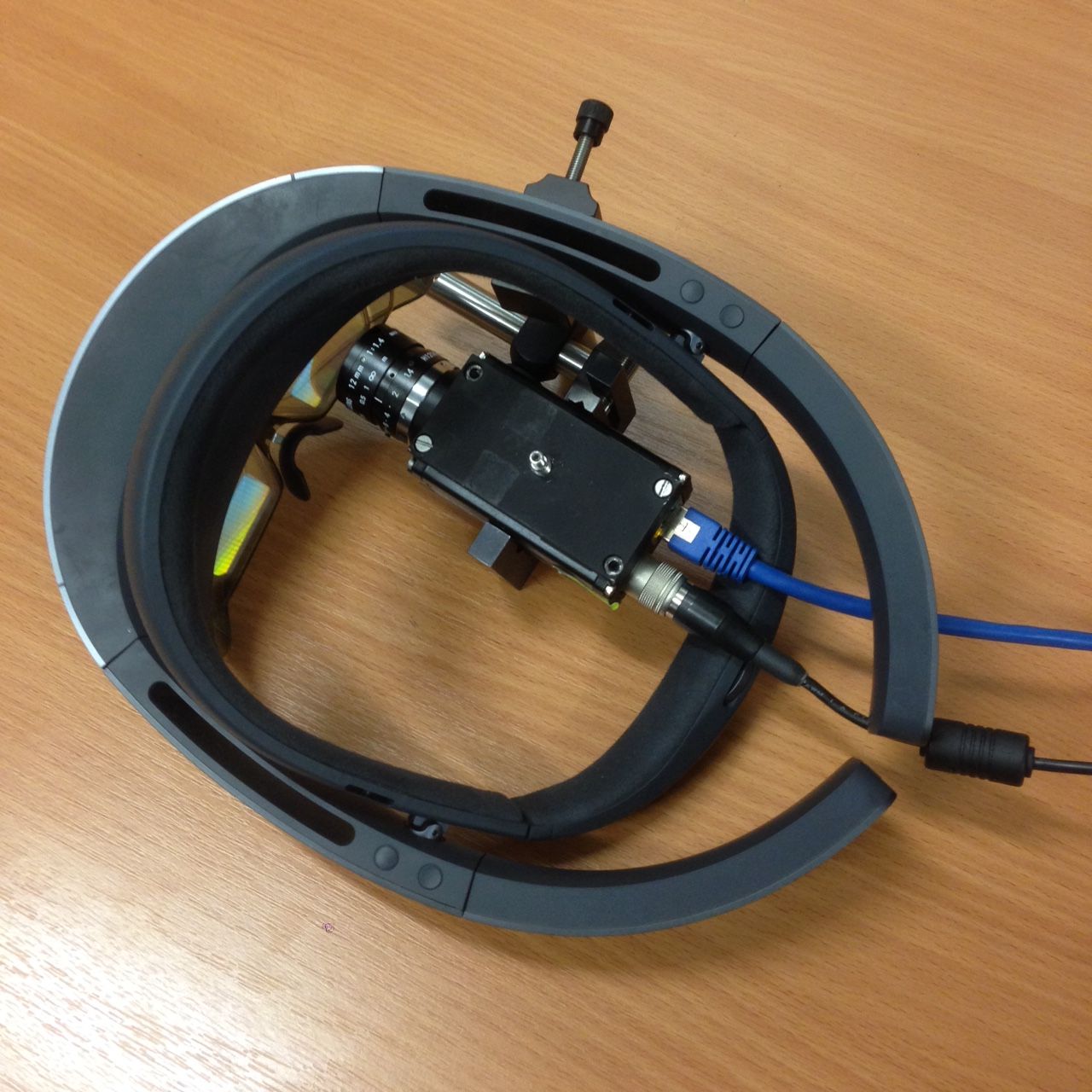}\\
\caption{A Basler Scout camera was attached to the HoloLens to capture materials for the Figures and supplementary videos.}
 \label{fig:headset}
\end{figure}

Examples of faces augmented with HoloFace are shown in Figures \ref{fig:example} and \ref{fig:sequence}. The images shown in the figures above were all taken directly through the display of the HoloLens using a camera attached to the headset, one of the setups we used is shown in Figure \ref{fig:headset}.

We publish\footnote{\url{https://github.com/MarekKowalski/HoloFace}} the entire code of HoloFace under the MIT License.

\subsection{Datasets and training}
We train both the KRFWS and DAN face alignment methods on the 300-W \cite{300W} and Menpo \cite{Menpo} datasets, with some additional images from the Multi-PIE \cite{MultiPIE} dataset for the KRFWS method.
For the DAN method we additionally blur some of the input images for training to increase robustness against motion blur. Since the speed of the KRFWS method depends on the number of tracked landmarks we only use the 51 internal landmarks out of the 68 contained in the datasets, rejecting the landmarks on the face edge. The methods used for tracking verification are trained on the 300-W \cite{300W} dataset for KRFWS and on the WIDER \cite{WIDER} dataset for DAN.

\subsection{Performance}
For both tracking methods the framerate of the pipeline during tracking is 30 fps, which is the maximal framerate of the built-in webcam, this includes the latency caused by the network traffic in the DAN face tracker. The framerate drops to 17 fps when no face is tracked and detection is being performed. In order to obtain a high processing speed with the local tracker, we have created a highly optimized implementation of the KRFWS method that runs at over 1000 fps on a desktop PC equipped with a 4 core CPU. The single stage DAN tracker runs at 160fps on a NVIDIA GeForce 1070GTX GPU (this is the execution time only, excluding the network transfer).

The high framerate of the remote face alignment method, DAN, is possible thanks to an efficient data transmission scheme. Instead of transmitting the entire image captured by the camera, only a small $112 \times 112$ image containing the face itself is sent. This particular size comes from the DAN method itself which uses this image size as input. Because of that the small image size does not have any adverse effect on the face alignment accuracy. Combined with auxiliary data the total bandwidth at 30 fps is only 3,1 Mbit/s.

The low bandwidth requirement potentially allows for using HoloFace with DAN tracking on GSM networks, using a mobile access point or, if future AR headsets allow it, a direct connection to the mobile network. This would allow a similar level mobility for the user with both the local and remote face tracking methods.

It is important to note that the framerate of HoloFace is independent of the framerate of the main thread of the application, which remains constantly at 60fps, which is the maximal value. This ensures a smooth AR experience and leaves room for additional load from other sources.

In terms of experience the two most visible issues of HoloFace are the delay caused by the acquisition time of the camera (more details in \ref{sec:delay}) and the noise produced by the local face tracker. The impact of both issues is significantly reduced by the filtering methods described in section \ref{sec:prediction}. For details on how HoloFace performs, please see the video in supplementary materials.

\section{Experiments}  \label{sec:experiments}
\subsection{HoloLens webcam delay}\label{sec:delay}
The webcam of Microsoft HoloLens, which is the only camera on the device currently available to the developers, has a significant delay between the moment the frame is acquired and the moment it is returned by the API. We have measured the delay using a method similar to the one proposed in \cite{MakeupLamps}. A high speed camera (we used a Basler ace series camera) is attached to the headset so that it sees the display. Subsequently the HoloLens is set to display the most recent image acquired by the camera with no other tasks running. Then the rig consisting of the HoloLens and the camera is pointed at a high speed clock. The difference between the time seen by the camera on the clock and the time shown on the image displayed by the headset (also recorded by the camera) is the time between image acquisition and rendering. In our experiments this time was $137\pm15$ms. Together with the processing time, the average delay between acquisition and rendering for the whole pipeline is $170$ms.

\subsection{Face tracking evaluation}\label{sec:tracking_accuracy}
\begin{figure*}
\centering
\includegraphics[width=0.33\textwidth]{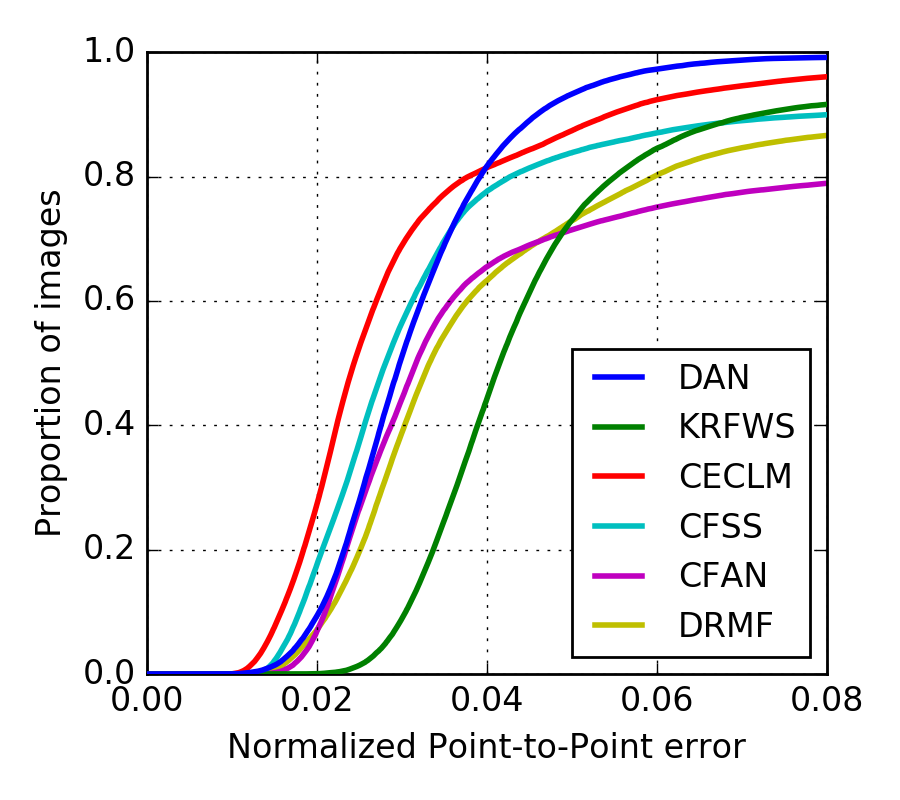}
\includegraphics[width=0.33\textwidth]{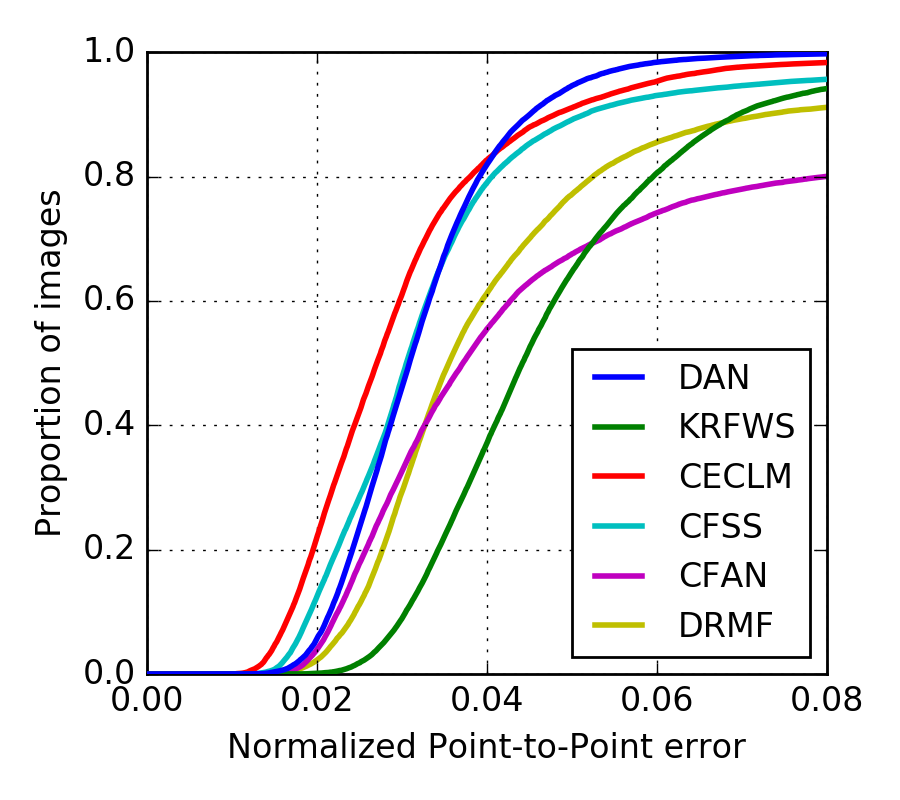}
\includegraphics[width=0.33\textwidth]{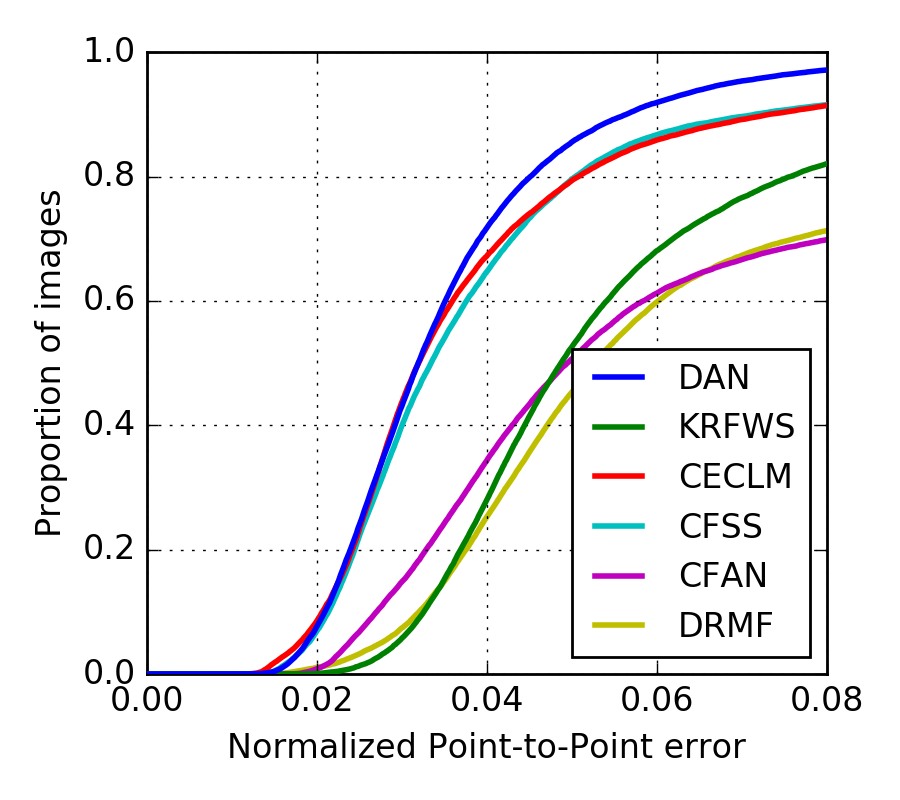}
\caption{Cumulative distribution curves of face tracking methods on the 300-VW \cite{300VW} dataset.}
 \label{fig:CED}
\end{figure*}

In order to evaluate the two face tracking methods, as well as the novel failure detection method used in DAN tracking, we perform experiments on the 300-VW \cite{300VW} face tracking dataset. The dataset consists of videos in three categories, where the third contains the most difficult sequences, more details in \cite{300VW}. In order to compare to the state of the art we use the results of an evaluation of several methods recently published in \cite{Zadeh}. Since the authors of \cite{Zadeh} only use the 49 internal landmarks we follow the same protocol and reject the 17 landmarks on the face's edge as well as the two inner mouth corners.

In each video we detect the face using OpenCV and track it using DAN and KRFWS until a loss of tracking is detected with the corresponding method, once that happens face detection is performed again. The experiments performed in \cite{Zadeh} use a different strategy, the face is detected in every 30th frame and tracked in between. There is no tracking failure detection and the face detection method is more potent \cite{MultiTask}. It also has to be mentioned that the methods used in HoloFace are trained on different datasets than the methods described in \cite{Zadeh}. 

Following \cite{Zadeh} we use the mean distance between the localized landmarks and the ground truth landmarks divided by the inter-ocular distance (the distance between the outer eye corners) as the error metric. This metric is also known as the normalized point-to-point distance. Based on this metric we plot the Cumulative Error Distribution (CED) curves shown in Figure \ref{fig:CED}. We also calculate the failure rate, which is the percentage of images with an error greater than $0.08$, and the area under the CED curve up to the threshold of $0.08$ ($AUC_{0.08}$). The results of the above experiments are shown in Table \ref{tab:results1} and \ref{tab:results2}.

Even though only a single stage is used, DAN combined with our failure detection method obtains the lowest failure rate on all the subsets and the best $AUC_{0.08}$ score on the hardest category. The KRFWS is significantly less precise, with low $AUC_{0.08}$ scores, the failure rate however is less than 10\% for the first two subsets. Even though the results are poorer we believe that the use of this method is justified thanks to its extremely high efficiency (over 1000 fps on a desktop computer as mentioned above). The results of KRFWS would definitely be improved if the number of stages, or the feature size were increased, this would however lower the framerate on the HoloLens to less than 30 fps, which we tried to avoid at all cost.  

\begin{table}[tb]
\caption{$AUC_{0.08}$ of the face tracking methods on the 300VW test set. } \label{tab:results1}

\begin{tabularx}{\linewidth}{ >{\centering\arraybackslash}X c c c}
\Xhline{4\arrayrulewidth}
Method & Category 1 & Category 2 & Category 3 \\
\hline
DRMF \cite{DRMF} & 48.70 & 48.75 & 30.18\\
CFAN \cite{CFAN} & 47.64 & 44.02 & 32.96\\ 
CFSS \cite{CFSS} & 56.88 & 57.83 & 51.62\\
CE-CLM \cite{Zadeh} & 62.53 & 62.28 & 52.33\\
\hline	
\textbf{KRFWS\footnotemark} & \textbf{43.22} & \textbf{41.06} & \textbf{33.98}\\
\addtocounter{footnote}{-1}
\textbf{DAN\footnotemark} & \textbf{60.05} & \textbf{59.53} & \textbf{55.44}\\
\Xhline{4\arrayrulewidth}
\end{tabularx}
\end{table}

\addtocounter{footnote}{-1}

\begin{table}[tb]
\caption{Failure rate percentage of the face tracking methods on the 300VW test set. } \label{tab:results2}
\begin{tabularx}{\linewidth}{ >{\centering\arraybackslash}X c c c}
\Xhline{4\arrayrulewidth}
Method & Category 1 & Category 2 & Category 3 \\
\hline
DRMF \cite{DRMF} & 13.38 & 8.86 & 28.71\\
CFAN \cite{CFAN} & 21.08 & 19.96 & 30.15\\ 
CFSS \cite{CFSS} & 10.64 & 4.34 & 8.43\\
CE-CLM \cite{Zadeh} & 3.94 & 1.65 & 8.55\\
\hline	
\textbf{KRFWS\footnotemark} & \textbf{8.39} & \textbf{5.85} & \textbf{17.92}\\
\addtocounter{footnote}{-1}
\textbf{DAN\footnotemark} & \textbf{0.83} & \textbf{0.25} & \textbf{2.86}\\

\Xhline{4\arrayrulewidth}
\end{tabularx}
\end{table}

\footnotetext{Note that those results are for the simplified, faster version of the method, which we use in this work, and are not the same as the results of the full method.}

\section{Conclusions}
We have presented HoloFace, an open-source framework for real time face alignment, head pose estimation and facial attribute retrieval for Microsoft HoloLens. With the help of our application Augmented Reality developers and researchers can add a new element of interaction into their projects. 

Some of the areas we believe are interesting for future work on HoloFace are: development of a new local face tracking method, which would be more accurate and less noisy, validation of face tracking using the remote method over a GSM network.

We also hope that future Windows Mixed Reality devices will give the developers access to the low latency cameras used by the headset for tracking. This would significantly improve the experience thanks to the shorter time between image acquisition and rendering.

\section{Acknowledgements}
We would like to thank the authors of \cite{Zadeh} for sharing the detailed results of their evaluation with us. We also thank the NVIDIA Corporation for donating a Titan X Pascal GPU which was used to train the Deep Alignment Network \cite{DAN} model used in this work.

{\small
\bibliographystyle{ieee}
\bibliography{egbib}
}

\end{document}